\documentclass[10pt,twocolumn,letterpaper]{article}

\usepackage{iccv}
\usepackage{times}
\usepackage{epsfig}
\usepackage{graphicx}
\usepackage{amsmath}
\usepackage{amssymb}
\usepackage{xcolor}
\usepackage{threeparttable}
\usepackage{subcaption}
\usepackage[ruled,vlined]{algorithm2e}

\newcommand{\FF}{\mathcal{F}}
\newcommand{\JJ}{\mathcal{J}}
\newcommand{\Jbf}{\mathbf{J}}
\newcommand{\Jbfh}{\mathbf{\hat{J}}}
\newcommand{\GG}{\mathcal{G}}
\newcommand{\Scal}{\mathcal{S}}
\newcommand{\LL}{\mathcal{L}}
\newcommand{\PP}{\mathcal{P}}
\newcommand{\Pbf}{\mathbf{P}}
\newcommand{\Pbfh}{\mathbf{\hat{P}}}
\newcommand{\Abf}{\mathbf{A}}
\newcommand{\Lbf}{\mathbf{L}}
\newcommand{\Lbfh}{\mathbf{\hat{L}}}

\newcommand{\R}{\mathbb{R}}
\newcommand{\RR}{\mathcal{R}}
\newcommand{\ff}{\mathbf{f}}

\newcommand{\gbf}{\mathbf{g}}
\newcommand{\gbfh}{\mathbf{\hat{g}}}
\newcommand{\hh}{\mathbf{h}}
\newcommand{\cc}{\mathbf{c}}
\newcommand{\Cbf}{\mathbf{C}}
\newcommand{\cch}{\mathbf{\hat{c}}}
\newcommand{\dd}{\mathbf{d}}
\newcommand{\ddh}{\mathbf{\hat{d}}}
\newcommand{\abf}{\mathbf{a}}
\newcommand{\zz}{\mathbf{z}}
\newcommand{\sbf}{\mathbf{s}}
\newcommand{\wbf}{\mathbf{w}}
\newcommand{\ZZ}{\mathbb{Z}}
\newcommand{\uu}{\mathbf{u}}
\newcommand{\xx}{\mathbf{x}}
\newcommand{\uuh}{\mathbf{\hat{u}}}
\newcommand{\vv}{\mathbf{v}}
\newcommand{\vvh}{\mathbf{\hat{v}}}
\newcommand{\pAP}{\text{pAP}}
\newcommand{\mpAP}{\text{mpAP}}

\newcommand{\figwidth}{0}
\newcommand{\imgname}{0}
\newcommand{\hatchsmall}{}
\newcommand{\hatchlarge}{}

\newcommand{\exampleImageSynthetic}[1]{%
\renewcommand{\imgname}{#1}%
\includegraphics[width=\figwidth]{examples/gt_hgt/\imgname_E00_all_lines_comp}
\includegraphics[width=\figwidth]{examples/gt_cycle_sampling_polygon/\imgname_E00_planes\hatchsmall_comp}
\includegraphics[width=\figwidth]{examples/gt_hgt/\imgname_E00_planes\hatchsmall_comp}
\includegraphics[width=\figwidth]{examples/gt_ann/\imgname_GT\hatchlarge_comp}
}
\newcommand{\exampleImageJoint}[1]{%
\renewcommand{\imgname}{#1}%
\includegraphics[width=\figwidth]{examples/det_hgt/\imgname_E00_all_lines_comp}
\includegraphics[width=\figwidth]{examples/det_hgt/\imgname_E00_planes\hatchsmall_comp}
}

\graphicspath{{./fig/}}

\usepackage[pagebackref=true,breaklinks=true,colorlinks,bookmarks=false]{hyperref}

\iccvfinalcopy 

\ificcvfinal\fi

\begin{document}

\title{Polygon Detection for Room Layout Estimation using \\ Heterogeneous Graphs and Wireframes}

\author{David Gillsjö \and Gabrielle Flood \and Kalle Åström\\
Centre for Mathematical Sciences\\
Lund University, Sweden\\
{\tt\small \{david.gillsjo,gabrielle.flood,kalle.astrom\}@math.lth.se}
}

\maketitle
\ificcvfinal\thispagestyle{empty}\fi

\begin{abstract}
   This paper presents a neural network based semantic plane detection method utilizing polygon representations. The method can for example be used to solve room layout estimations tasks. The method is built on, combines and further develops several different modules from previous research. The network takes an RGB image and estimates a wireframe as well as a feature space using an hourglass backbone. From these, line and junction features are sampled. The lines and junctions are then represented as an undirected graph, from which polygon representations of the sought planes are obtained. Two different methods for this last step are investigated, where the most promising method is built on a heterogeneous graph transformer. The final output is in all cases a projection of the semantic planes in 2D. The methods are evaluated on the Structured 3D dataset and we investigate the performance both using sampled and estimated wireframes.
   The experiments show the potential of the graph-based method by outperforming state of the art methods in Room Layout estimation in the 2D metrics
   using synthetic wireframe detections.
\end{abstract}
%
\section{Introduction}
\begin{figure}
\begin{center}
\includegraphics[width=\linewidth]{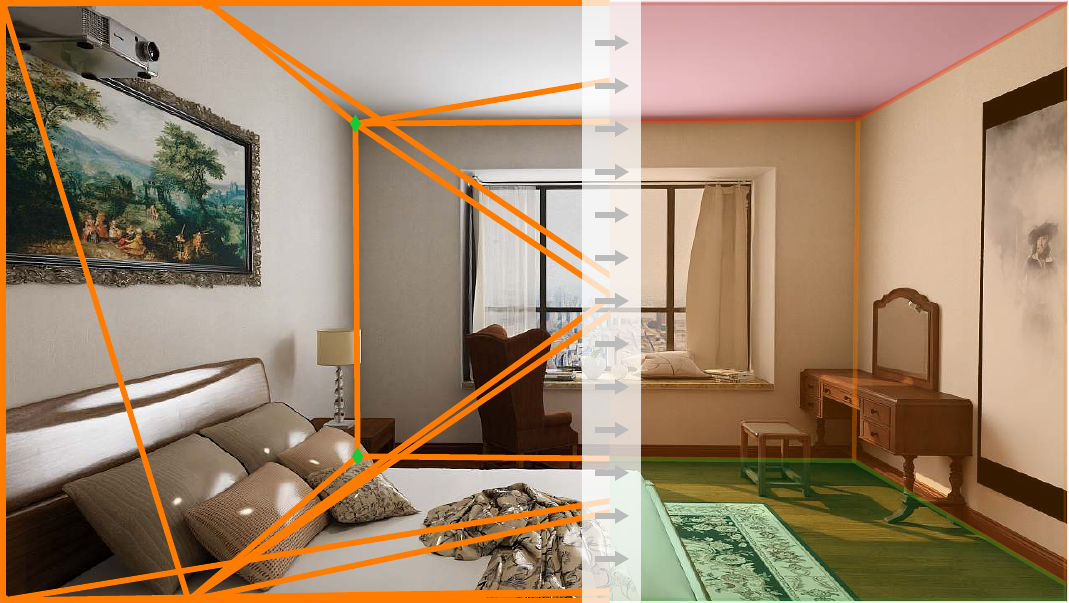}
\end{center}
\caption{The model takes an image and a wireframe to produce a set of polygons corresponding to a Room Layout.}
\vspace{-2mm}
\label{fig:motivation}
\end{figure}
With all the advanced methods in Computer Vision today, it is possible to perform Structure from Motion and 3D reconstruction in many ways.
Depending on your application a generic point-based method might work well, but in other cases it may be beneficial to model projections as other geometric objects.

In this paper we present a method for detecting the projections of planes in an image, where each projection may be represented as a polygon.
The input to the suggested system is an RGB image and the first step in the pipeline is to detect a wireframe consisting of line segments in this image,
as shown in Figure \ref{fig:motivation}. The line segments form a wireframe which together with different types of features gathered from a backbone are used to create a graph representation of the scene. A set of polygons is then obtained from this graph using either a cycle generating method or a Heterogeneous Graph Transformer (HGT) network. Furthermore, features connected to these polygons are then obtained either using sampling or further networks. Finally, a plane feature classifier is used to conclude which polygons that actually represent planes in the image, and the semantic class of these polygon planes. An overview of the developed systems is shown in Figure~\ref{fig:variations}.

The experiments in this paper focus on the Room Layout Estimation task, where each wall, floor and ceiling is represented as a plane polygon in 3D. Even though our system does not produce any depth information, we can still compare the 2D projections of the room layout.
We show that given good enough line segment detections, the model successfully predicts a Room Layout projection with performance matching the state of the art models.
While there certainly are other tasks that fit the assumptions made with this model, there are good datasets for Room Layout Estimation to use when training the model. This means that our plane polygon detector will be evaluated on this type of tasks, but the usage is not limited to this and may be extended in the future.

Previous work in Room Layout Estimation that use Neural Network (NN) models produce an intermediate representation that is later on
used in an heuristic optimization method to generate the Room Layout.
For example Lin \etal \cite{lin2018layoutestimation} predict a semantic segmentation which is later used to optimize intersection lines heuristically.
Other methods \cite{cfile,zhao2017physics,Yan,doubleref2020} predict edges and then optimize.
The most recent work of \cite{NonCuboidRoom2022} use a combination of plane, depth and vertical line detections to heuristically fuse and generate a Room Layout.
Most of these methods also place strong priors on the room shape, for example requiring cuboid shape or fixed angles between planes.
We present a method in which the NN model reasons jointly over line predictions to directly estimate a Room Layout with the only constraint being
that the room is a set of intersecting planes. To the best of our knowledge, this is the first end-to-end NN model to directly output a Room Layout without cuboid requirements.


\noindent
\textbf{The contributions of this paper are:}
\begin{itemize}
	\item A novel Heterogeneous Graph Neural Network (HGNN) model \footnote{\url{https://github.com/DavidGillsjo/polygon-HGT}} and a polygon sampling model for estimating Room Layouts from line detections. The model do not require any heuristics in post-processing.
	\item Results on the dataset Structured3D using generated and learned line proposals. This study gives insights about the importance of the graph structure of the problem.
	\item A quantitative comparison with state of the art models in Room Layout estimation.
\end{itemize}

\section{Related work}
\textbf{Plane Instance Detection} is a related task where the goal is to detect each plane in the image and provide a pixel mask and 3D parameter estimate for each plane instance with the camera as origin. The planar objects are foreground objects so this task does not deal with occlusions, contrary to Room Layout estimation.
The state of the art models are all NN models \cite{plane_yu2019, plane_yang2018, Liu_2019_CVPR, liu_2018_cvpr, tan2021planeTR} with impressive performance on instance level.

\textbf{Wireframe Parsing.} As input for our model we use line segment detections from wireframe parsing \cite{xia2014accurate,xue2019learningcvpr,xue2019learningpami, HAWP}.
These methods predict connected line segments without any semantic meaning. The most common pipeline is to start from junction proposals and then
work toward line segments, but some works like \cite{lgnn2020} directly use line predictions together with a Graph Neural Network.
Recently, Gillsj{\"o} \etal \cite{srw-net} predicts a semantic wireframe based on room geometry using a Graph Convoluional Network, but this does not account for plane instances.

\textbf{Room Layout Estimation} has been studied in many forms. Hedau \etal \cite{Hedau} early on used the Manhattan World assumption \cite{manhattan} as prior for the room shape.
Prior to NN models \cite{prenetwork1,prenetwork2,prenetwork3} used handcrafted features as a base, then did vanishing point detection and hypothesis generation.
Recent approaches use CNNs, \eg Mallya and Lazebnik \cite{Mallya} utilize structured edge detection forests with a CNN to predict an edge probability mask.
Lin \etal \cite{lin2018layoutestimation} proposed an end-to-end CNN for pixelwise segmentation of the room image with post processing.
DeepRoom3D \cite{deeproom} use an end-to-end CNN to predict a cuboid and RoomNet \cite{lee2017roomnet} directly predicts ordered keypoints in a room layout.
A group of recent methods \cite{cfile,zhao2017physics,Yan,doubleref2020} use CNNs to predict edges and then optimize for the Room Layout using geometric priors.

All these approaches have strong priors on the Room geometry. There have recently been some methods considering a general model without these constraints.
Stekovic \etal \cite{general3d2020} solve a discrete optimization problem over 3D polygons using both RGB and Depth information.
Howard-Jenkins \etal \cite{HowardJenkins2018ThinkingOT} use plane detection to form a 3D model over a video sequence.
Yang \etal \cite{NonCuboidRoom2022} use a combination of plane, depth and vertical line detections to estimate a general Room Layout.
The datasets mostly used are LSUN \cite{yu2016lsun} and Hedau \cite{Hedau}. In this paper we, similarly to \cite{NonCuboidRoom2022}, will use Structured3D \cite{Structured3D}.
There is also a lot of work done in Room Layout estimation for panorama images, but as this work focuses on perspective images we deem it to be outside of the scope.


\section{Model}
\begin{figure}[t]
\begin{center}
\includegraphics[width=0.8\linewidth]{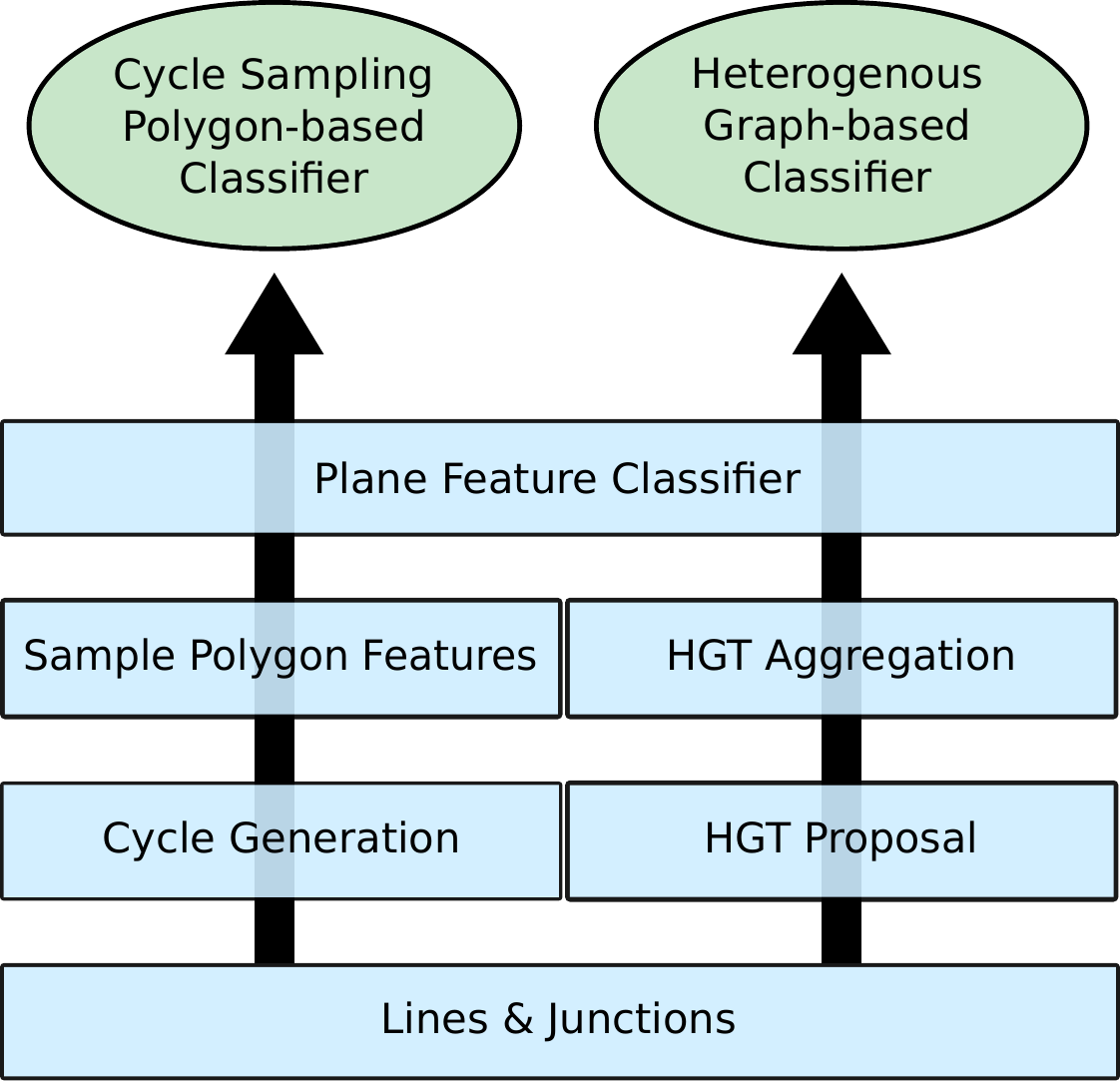}
\end{center}
\caption{Two different architectures are evaluated. The Cycle Sampling Polygon-based Classifier generate cycles heuristically from input lines
while the The Heterogeneous Graph-based Classifier uses a Neural Network on Heterogeneous graphs.}
\label{fig:variations}
\end{figure}

In this paper we evaluate two different models for finding semantic planes in images, which both share the same backbone. The models will be described in detail later but here, and in Figure \ref{fig:variations}, we give an overview of the two different models.
The Cycle Sampling Polygon-based Classifier (CSP) generates possible polygons from detected line segments and samples features from the polygon for classification.
The Heterogeneous Graph-based Classifier (HGC) predicts plane centroids to use as anchors. These anchors are together with line and junction features encoded into a
heterogeneous graph which is fed through the network. In the end, each plane node is classified using an MLP classifier.

\subsection{Backbone}
The backbone is an Hourglass Neural Network \cite{hourglass} which takes a resized RGB image of size $3 \times 512 \times 512$ pixels as input and outputs a
feature space $\FF$ with dimensions $128 \times 128 \times 256$, where 128 is spatial size and 256 is the number of channels.
This feature space is used for sampling both junction, line and plane features.

\subsection{Line and Junction Feature Sampling} \label{sec:jl_sampling}
The method for line and junction feature sampling is based on HAWP \cite{HAWP}, but we add labels to the junctions to predict whether
the junctions are corresponding to actual junctions in 3D or end a line segment due to occlusions.
Similarly to SRW-net \cite{srw-net}, if a junction has a 3D correspondence we call it \textit{proper}, otherwise \textit{false}.

Let $J_i = \{\zz_i, \ff_i\}$ denote a junction with index $i$ and $\Jbf=\{J_1,J_2,\hdots,J_R\}$ a set of junctions.
Each junction consists of the coordinates $\zz_i \in \R^2_{[0,128)}$
and a feature vector $\ff_i \in \R^{256}$. This feature vector is generated by
\begin{equation}
  \ff_i = \JJ(\FF(\lfloor \zz_i \rfloor)),
\end{equation}
where $\JJ$ is $\text{Conv2D}(256,3) \rightarrow \text{ReLU}$.
The notation for the convolution layer is $\text{Conv2d}(\text{output channels}, \text{kernel size})$.

Similarly, let $L_j = \{\uu_j, \vv_j, \gbf_j\}$ denote a line segment with index $j$ consisting of end point coordinates $(\uu_j,\vv_j) \in \R^2_{[0,128)}$
and a feature vector $\gbf_j \in \R^{512}$. The endpoints are also junctions, and the feature vector is generated by a network head $\LL_{NN}$ and a sampling function $\LL_s$
where $\LL_{NN}$ is $\text{Conv2D}(128,3) \rightarrow \text{ReLU}$.
$\LL_s$ takes 16 sample points uniformly spread between endpoints $(\uu_j,\vv_j)$  and interpolates a feature vector of size $16 \times 128$,
this is down sampled by max pooling to $4 \times 128$ and flattened so that $\LL_S(L_j) \in \R^{512}$.
In summary the vector $\gbf_j$ is formed by taking
\begin{equation}
  \gbf_j = \LL_s(\LL_{NN}(\FF (\uu_j, \vv_j))).
\end{equation}

%
%

\subsection{Cycle Sampling Polygon-based Classifier} \label{sec:CSP}

The first architecture (CSP) generates possible polygons from
detected line segments and samples features from the polygon for classification.
Given a set of connected line segments $\Lbf = \{L_1, L_2, ..., L_M \}$ we form an undirected graph $\GG$
with endpoints as vertices and form an edge for every detected line segment.
From $\GG$ all possible polygons may be generated. This is done by generating cycles using the cycle basis of $\GG$ \cite{lee1982algorithmic}.
A cycle is a path of connected vertices $V = \{v_1, v_2, ..., v_n, v_1\}$ where only the first and last vertex are equal.
All simple cycles are not polygons, so we require each cycle to have at least three unique vertices
and the generated polygon to have no intersecting boundaries.

\begin{figure}[t]
\begin{center}
\includegraphics[width=0.3\linewidth]{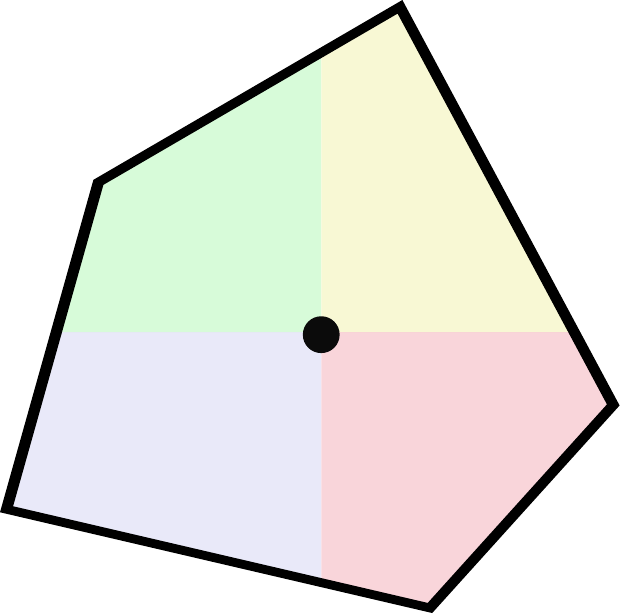}
\end{center}
\caption{To sample features from within the polygon it is split in four quadrants with the centroid as origin.
Features from all pixels in each quadrant are max pooled.}
\label{fig:polygon_sampling}
\end{figure}

This gives a set of polygons $\Pbf = \{ P_1, P_2, ..., P_K \}$.
The Cycle Sampling Polygon-based Classifier proceeds to sample features from inside the polygon, similar to how object detectors like Mask-RCNN \cite{maskrcnn}
samples from inside an object's bounding box. The sampling is done in three parts, first \textit{(i)} all pixels of $\FF$ are processed by
a CNN layer $\PP_{CNN}$ consisting of $\text{Conv2D}(128,3) \rightarrow \text{ReLU}$. Then \textit{(ii)} for each polygon $P_k$ its centroid $\cc_k \in \R^2_{[0,128)}$ is
calculated. With $\cc_k$ as origin $P_k$ is divided into four quadrants, see Figure \ref{fig:polygon_sampling}, where $\PP_{CNN}(\FF)$ for each quadrant is max pooled.
This yields a feature vector of size $4 \times 128$ which is flattened to be the polygon feature $\hh_k \in \R^{512}$.
The feature vector $\hh_k$ is then \textit{(iii)} put through a 3 layer MLP for classification.

A major downside of the cycle sampling approach is that the number of possible cycles may grow exponentially with the number of edges and vertices.
During inference it is also required to find all possible cycles to ensure that any valid polygon is found.
However, at training time it is not necessary to find all cycles, so to speed up training only 20 polygons are sampled.
We also add 10 positive examples and 10 negative examples of polygons from the annotations to stabilize training.

\subsection{Heterogeneous Graph-based Classifier}\label{sec:HGC}
\begin{figure}[t]
\begin{center}
\def\svgwidth{\linewidth}
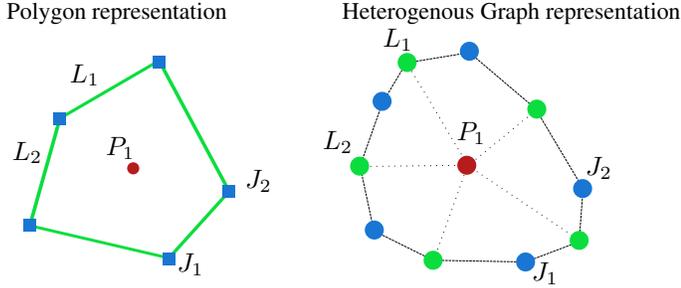
\end{center}
\caption{Illustration of how each polygon (left) may be represented with a heterogeneous graph (right) consisting of
junction, line and plane vertices. In this illustration there are two types of edges, junction-line and line-plane.}
\label{fig:heterogeneous_graph}
\end{figure}

The second architecture (HGC) is based on the Heterogeneous Graph Transformer 
\cite{hgt}.
An heterogeneous graph may consist of vertices and edges of different types.
We naturally use the node types junction, line and plane, and also different edge types in the proposal network and the classifier network.
Apart from the feature vectors for each entity we also encode the geometric position for the junctions, the midpoint for line segments, and the centroid for plane vertices.
Figure \ref{fig:heterogeneous_graph} shows an illustration on how we represent a polygon as a heterogeneous graph.

\begin{figure*}
\begin{center}
\includegraphics[width=\linewidth]{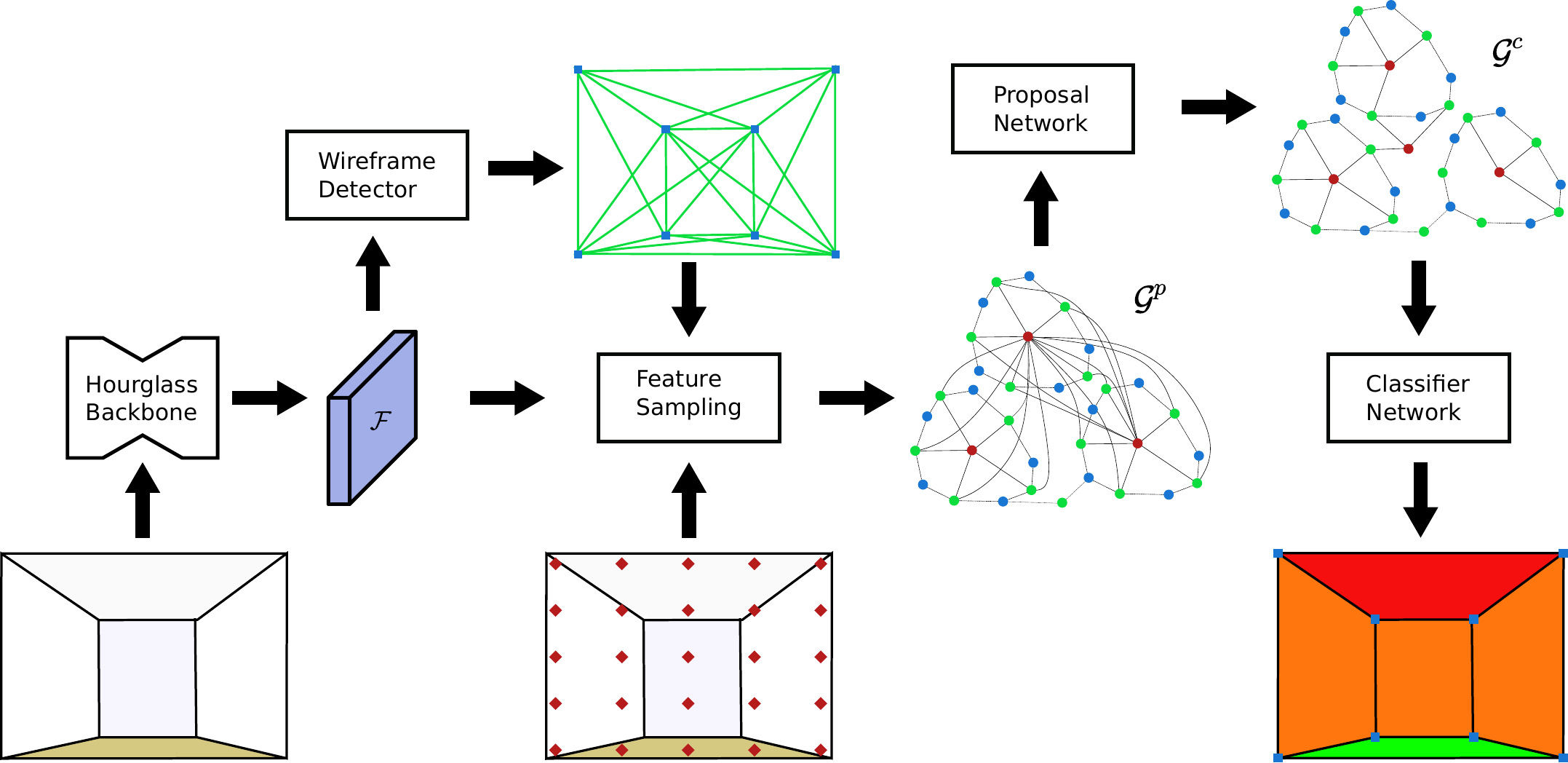}
\end{center}
\caption{Simplified illustration of the Heterogeneous Graph-based Classifier architecture. The image is fed through a backbone to yield a feature space $\FF$.
The Wireframe input is then used together with plane centroid anchors to sample features from $\FF$.
These are passed to the proposal network and the classifier network as heterogeneous graphs.}
\label{fig:architecture}
\end{figure*}

An overview of the full architecture of the Heterogeneous Graph-based Classifier (HGC) can be seen in Figure \ref{fig:architecture}.
The junction and line features are sampled from $\FF$ according to Section \ref{sec:jl_sampling}, and geometric information is concatenated with the features.
The proposal network takes the junctions $\Jbf$ and lines $\Lbf$ and generate proposal polygons $\Pbf$.
The polygons are together with the lines and junctions passed to the classifier network which labels each polygon.

\subsubsection{Heterogeneous Graph Transformer}
To jointly reason about planes, lines and junctions we use the Heterogeneous Graph Transformer (HGT) model \cite{hgt}
to perform message passing on the heterogeneous graph. Like a convolution operation in a CNN the message passing
updates the current node features with a weighted sum of nearby node features. Instead of specifying a kernel we may choose a hop length.
If the hop length is one, then all direct neighbours of node $t$ are source nodes $N[t]$ for the message passing.
A standard model to attention-based message passing in a homogeneous GNN for calculating layer $b$ is
\begin{equation}
H^b[t] = \underset{\forall s \in N[t]}{\text{Aggregate}} \left (
\text{Message} (s) \cdot \text{Attention}(s,t)
\right ),
\end{equation}
where $\text{Message}(s)$ in the simplest case is $H^{b-1}[s]$, a neighbour node feature from previous layers.
The $\text{Aggregate}$ operation may be a sum, mean or other operation that suits the application.

For a heterogeneous graph each node $t$ may have a different type $\tau(t)$ and each edge $e$ a different edge type $\theta(e)$.
The HGT model calculate different attention weights based on the meta relation $\langle \tau(t), \theta(e), \tau(s) \rangle$.
For more details on the model, including aggregation and message construction, see the paper by Hu \etal \cite{hgt}.

\subsubsection{Proposal Network}\label{sec:proposal_network}
Let $\Pbf_A = \{A_1, A_2, ..., A_{25} \}$ denote a set of plane anchors placed uniformly on a $5 \times 5$ grid.
Plane anchor $A_k$ has coordinates $\abf_k \in \ZZ^2_{[0,127]}$ and features $\dd_k$ generated by
\begin{equation}
  \dd_k = \PP_{p}( \PP_{CNN}(\FF), \abf_k),
\end{equation}
where $\PP_{CNN}$ is the same as for CSP in Section \ref{sec:CSP} and $\PP_{p}$ is a multi-scale convolution operation
with four $\text{Conv2D(32,h,d)}$ operations of kernel size $h=1,3,3,3$ and dilation $d=0,0,1,2$, respectively. The output of these four are concatenated
to form the output of $\PP_{p}$.

\begin{figure*}
\begin{center}
\def\svgwidth{\linewidth}
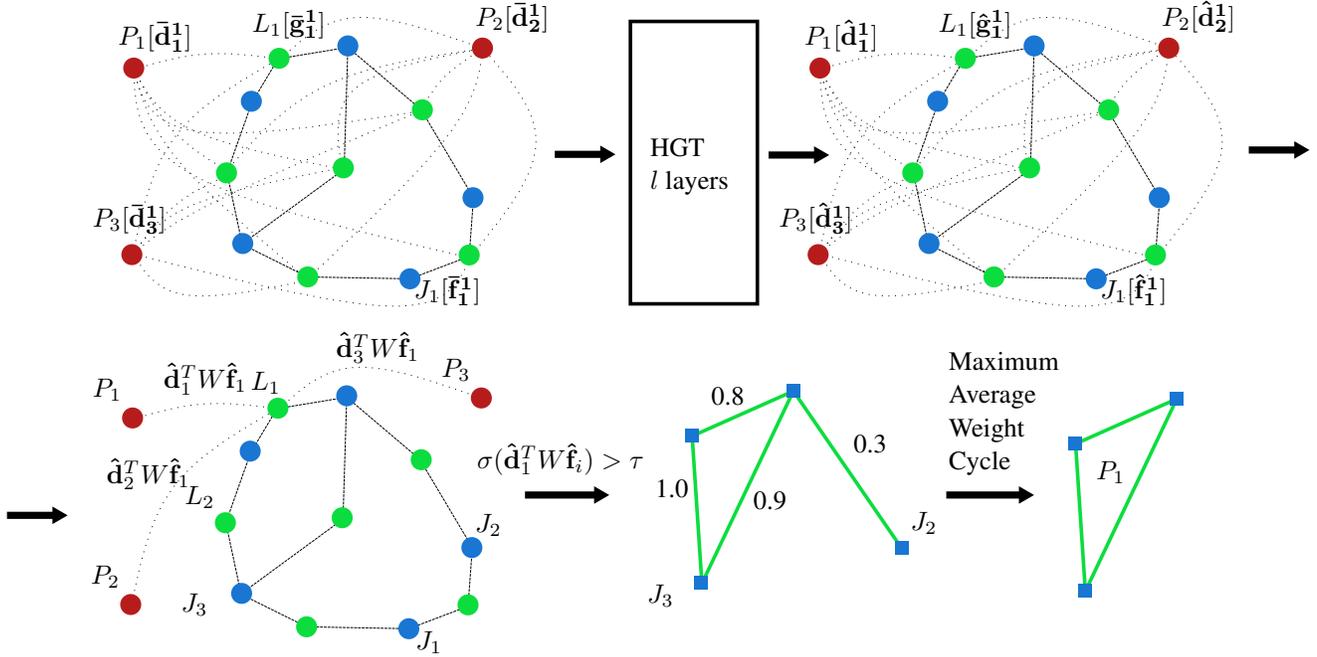
\end{center}
\caption{Illustration of the Proposal network in the Heterogeneous Graph-based Classifier. (i) All plane nodes are linked with all line nodes prior in the graph.
(ii) The graph is fed to the HGT model and new features are computed. (iii) Link prediction between line and plane nodes is performed,
edges with score $>\tau$ are transformed to a homogeneous graph and (iv) a polygon is found by finding the cycle with maximum average weight.}
\label{fig:link_pred}
\end{figure*}

Now the graph $\GG^p$ for HGT is constructed. The features are
\begin{align}
  \bar{\ff}_i  &= \begin{bmatrix} \bar{\zz}_i & \ff_i \end{bmatrix},                          \quad &&\text{for} \quad J_i \in \Jbf , \label{eq:junction_feature}  \\
  \bar{\gbf}_j &= \begin{bmatrix} \frac{\bar{\uu}_j + \bar{\vv}_j}{2} & \gbf_j \end{bmatrix}, \quad &&\text{for} \quad L_j \in \Lbf ,   \label{eq:line_feature} \\
  \bar{\dd}_k  &= \begin{bmatrix} \bar{\abf}_k & \dd_k \end{bmatrix},                         \quad &&\text{for} \quad A_k \in \Pbf_A ,
\end{align}
where $\bar{\zz}=\frac{\zz}{128}$ to normalize the coordinates and likewise for $\bar{\uu},\bar{\vv},\bar{\abf}$.
As illustrated in Figure \ref{fig:link_pred} the graph is passed through an HGT of $l$ layers which yields a new graph $\GG^p_l$ with the same
structure but updated features $\hat{\ff_i},\hat{\gbf_j},\hat{\dd_k}$. The graph $\GG^p_l$ is used to generate a plane proposal for each plane node.
This problem is formulated as a link prediction task, where we for each plane anchor $A_k$ predict edges to line nodes $L_j$.
For each pair $j,k$ a score is predicted as $\sigma (\hat{\gbf_j^T} W \hat{\dd_k})$, where $\sigma(x) = \frac{1}{1+e^{-x}}$ is the sigmoid function
and $W$ is a weight matrix learned during training.
Then for each $A_k$ a homogeneous graph with junctions as nodes and lines as edges is formed using every line $j$ s.t $\sigma (\hat{\gbf_j^T} W \hat{\dd_k}) > \kappa$.
The edge weight correspond to how likely it is that the line should be in the plane with anchor $A_k$.
We find the cycle with Maximum average weight using an approximate greedy solution and take it as plane proposal $P_k$.

To train the network a binary cross entropy loss is used with bounded biparte matching.
First a biparte matching $\rho : \ZZ^+ \rightarrow \ZZ^+$ is found between all annotated line segments $\hat{L}_q \in \hat{\Lbf}$ and detected line segment $L_j \in \Lbf$ s.t.
\begin{align*}
  \rho^* &= \underset{\rho}{\text{argmin}} \; \sum_q  D(L_{\rho(q)}, \hat{L}_q), \\
  D(L_j, \hat{L}_q) &= \min  ( \| \uu_j-\uuh_q \|^2 + \| \vv_j-\vvh_q \|^2, \\
  & \| \uu_j-\vvh_q \|^2 + \| \vv_j-\uuh_q \|^2  )
\end{align*}
if $ D(L_{\rho(q)}, \hat{L}_q) < \alpha$, otherwise the match is excluded.

Let $\Pbfh$ denote the set of $G$ annotated ground truth planes with plane instances $ \hat{P}_m \in \Pbfh$ s.t. $\hat{P}_m = \{ \cch_m, \Lbfh_m \}$,
where $\Lbfh_m \subset \Lbfh$ is the set of $M_m$ line segments defining the polygon and $\cch_m$ the centroid of the polygon.
Another biparte mapping $\phi : \ZZ^+ \rightarrow \ZZ^+$ is made between annotated planes and plane anchors $A_k \in \Abf$ s.t.\ the distance between centroids are minimized, i.e.
\begin{equation}
  \phi^* = \underset{\phi}{\text{argmin}} \; \sum_m  \|\abf_{\phi(m)} - \cch_m \|.
\end{equation}

Now we may construct the loss function as
\begin{align*}
  \RR_{prop} &= \frac{1}{G} \sum_{m=1}^G \frac{1}{M_m} \left [
  \sum_{L_j \in \Phi_m} \log \sigma \left ( \gbfh_j^T W \ddh_{\phi^*(m)} \right ) \right.  \\
   &+ \sum_{L_j \notin \Phi_m} \log \left (1 - \sigma \left ( \gbfh_j^T W \ddh_{\phi^*(m)} \right ) \right)
  \left.\vphantom{\sum_{\hat{P}_m}}\right ],
\end{align*}
where $\Phi_m$ are the detected lines matched with lines in the matched annotated plane $\hat{P}_m$, i.e. $\Phi_m = \{ L_j \; | \;  \rho^*(q) = j \; \text{and} \; \hat{L}_q \in \hat{P}_m \}$.

\subsubsection{Classifier Network}
The classifier network takes the plane proposals $\Pbf = \{ P_1, P_2, ..., P_K \}$ and constructs a graph $\GG^c$
by taking previous junction and line features from Equation (\ref{eq:junction_feature}) and (\ref{eq:line_feature})
together with their structure from $\GG^p$. In addition plane features $\dd^c_k$ are sampled from the plane proposal polygons
by the same procedure as in the CSP model in Section \ref{sec:CSP}.
Just like in the proposal network, we add a normalized centroid coordinate $\bar{\cc}_k$ calculated from each polygon forming the final feature vector
\begin{equation}
  \bar{\dd}^c_k  = \begin{bmatrix} \bar{\cc}_k & \dd^c_k \end{bmatrix}, \quad \text{for} \quad P_k \in \Pbf.
\end{equation}
The graph $\GG^c$ is processed by a classifier HGT of $l$ layers yielding a new graph $\GG_l^c$ with plane features $\hat{\dd}^c_k$ for $P_k \in \Pbf$.
Final classification scores $\sbf \in \R^4_{[0,1]}$ for background, wall, floor and ceiling are given by network layers,
\begin{equation}
  \sbf_k = \text{Softmax}(\text{MLP}(\hat{\dd}^c_k)).
\end{equation}
The loss function is a cross entropy loss from bounded biparte matching. Using previous notation we find the mapping
\begin{equation}
  \phi^*_c = \underset{\phi_c}{\text{argmin}} \; \sum_m  \|\cc_{\phi_c(m)} - \cch_m \|,
\end{equation}
if  $\|\cc_{\phi_c(m)} - \cch_m \| < \tau_c$.
The loss function for annotated labels $y_m$ is
\begin{equation}
  \RR_{cl} = \frac{1}{G} \sum_{m=1}^G -\wbf[y_m] \; \sbf_{\phi^*_c(m)}[y_m],
\end{equation}
where $\wbf[y]$ is weight for class $y$ and $\sbf_k[y]$ is what $P_k$ scored for class $y$.

\subsubsection{Joint Wireframe estimation}
This model is also trained to jointly output a Wireframe detection from the same backbone.
The loss function $\RR_{wire}$ is defined by HAWP and so the total loss function for the network is
\begin{equation}
  \RR = \RR_{wire} + \RR_{prop} + \RR_{cl},
\end{equation}
with weights for each term used during training.

\subsubsection{Non-Maximum Suppression}
Though the graph model seems to do just fine without a Non-Maximum Suppression (NMS) we implement one for the output plane polygons.
A detected polygon $P_k \in \Pbf$ is suppressed if there is another polygon $P_r \in \Pbf$ with higher score $s_r > s_k$ and $\text{IoU}(P_r,P_k) > 0.05$.

\section{Evaluation}
To evaluate the models we cannot use all the Room Layout metrics since the models does not output depth information.
We do use the Intersection over Union (IoU) error and Pixel error from Stekovic \etal \cite{general3d2020}.
For these metrics we find a one-to-one correspondance $\theta : \ZZ^+ \rightarrow \ZZ^+$
mapping between the $K$ predicted plane polygons $\Pbf$ and the $M$ ground truth polygons $\Pbfh$ s.t
$\theta(m)=k$ iff. $\text{IoU}(\hat{P}_m, P_k) \geq \text{IoU}(\hat{P}_m, P_i)$ for $i \neq k$ starting from the largest ground truth polygon.
Given this mapping we calculate
\begin{align}
  \epsilon_I[{\text{IoU}}] &= \frac{2}{M+K} \sum_{m=1}^M \text{IoU}(\hat{P}_m, P_{\theta(m)}) , \\
  \epsilon_I[{\text{PE}}]  &= \frac{1}{|I|} \sum_{\xx \in I} \text{PE}(\xx) ,
\end{align}
where $\text{PE}(\xx)=0$ if the pixel is incorrectly matched and $\text{PE}(\xx)=1$ if correct and $|I|$ the number of pixels.
Furthermore, $\epsilon_I[\text{IoU}]$ and $\epsilon_I[\text{PE}]$ are calculated for each image $I$ and then averaged to form
$\epsilon[\text{IoU}]$ and $\epsilon[\text{PE}]$, respectively.

To measure performance in terms of semantic labeling and precision we use polygon Average Precision, $\pAP$,
similarly defined as AP metrics for object detection. It is the area under the precision recall curve.
A detection $P_k$ with label $q_k$ is a true positive if there is a ground truth polygon $\hat{P}_m$ with label $y_m=q_k$ and $\text{IoU}(\hat{P}_m, P_k) < \gamma$.
Only one detection may be matched to each ground truth polygon, extra detections are labeled false positives.
For each label we calculate $\pAP^\gamma$ for $\gamma \in [0.5,0.95]$ with step length $0.05$.
Then $\pAP^m = \underset{\gamma}{\text{mean}}(\pAP^\gamma)$ and we average over labels to get $\mpAP^m$.

\section{Experiments}
The models are trained and evaluated on the Structured3D dataset \cite{Structured3D} which is a large-scale, photo-realistic simulated dataset with 3D structure annotations.
It consists of 3'500 scenes, with a total of 21'835 rooms and 196'515 frames and has a predefined training, validation and test data split.
We use the existing annotations for the perspective images which include information about the planes such as polygon, 3D parameters and semantic label.
We evaluate performance between the three proposed models on simulated wireframe detections and compare the HGC model with
state of the art Room Layout Estimation when using both simulated wireframe detections and HAWP detections.

\subsection{Semantic Plane Detection on Room Layout}
In this experiment we use two different approaches to the Wireframe detector.
The most frequently used is a synthetic detector where line segments are generated from the ground truth annotations.
Given an image $I_i$ with $M_i$ line segments $\Lbfh_i$ and junctions $\Jbfh_i$ we generate line segments
$\Lbf^G_i = \{ L^G_1, L^G_2, ..., L^G_B \}$ according to L-CNN \cite{zhou2019end}. We generate to $B=M_i$ for each image $I_i$.
For the HGC model we also train a Wireframe detector jointly with the network. This is more challenging which is indicated by the footnote in Table \ref{tab:gen_AP}.
From the table we see that the HGC model is superior to the cycle sampling based model and is not as dependent on the NMS post-processing for performance.
The HGC performance on synthetic detections shows the potential given a correctly tuned Wireframe detector.
Performance declines when jointly predicting the Wireframe and the Polygons, likely due to worse wireframes and less flexibility in the model.
It is also clear the floor is the most difficult type of polygon, followed by wall and finally ceiling. Likely due to occlusions.




\begin{table}
	\begin{threeparttable}[b]
	  \caption{AP Scores for the 3 different models on generated data and joint prediction.}
	  \label{tab:gen_AP}
	  \centering
	  \begin{tabular}{|l|r|r|r|r|r|}
	  \hline
	   Model           & NMS         & $\pAP^m$ & $\pAP^m$ & $\pAP^m$ & $\mpAP^m$ \\
		 \hline
	                   &             &     wall &    floor &   ceiling &           \\
	  \hline
		CSP              & \checkmark  &      14.5 &     15.8 &      20.3 &      16.8 \\
	  CSP              &             &       4.6 &     4.2 &      5.6 &        4.8 \\
	  HGC              & \checkmark  &      93.6 &     90.5 &      97.6 &      93.9 \\
	  HGC              &             &      93.4 &     88.9 &      97.5 &      93.3 \\
		\hline
		HGC \tnote{+}    & \checkmark  &      55.8 &     39.6 &      65.9 &      53.8 \\
	  HGC \tnote{+}    &             &      55.1 &     40.4 &      65.6 &      53.7 \\
	  \hline
	  \end{tabular}
		\begin{tablenotes}
       \item [+] Joint Wireframe Detection.
    \end{tablenotes}
	 \end{threeparttable}
\end{table}

\subsection{Room Layout from Wireframe Detections}
Furthermore, we compare the best HGC models from both synthetic and joint wireframe detections with state of the art models for Room Layout estimation.
From Table \ref{tab:gen_room} we see that the HGC model with synthetic detections outperform the state of the art models by a large margin for the 2D metrics
$\epsilon[\text{IoU}]$ and $\epsilon[\text{PE}]$. However the HGC with joint wireframe detection cannot reach the same performance as competing models.

\begin{table}
	\begin{threeparttable}[b]
	  \caption{Room Layout scores for different models on generated data and joint prediction compared with state of the art models.}
	  \label{tab:gen_room}
	  \centering
	  \begin{tabular}{|l|r|r|}
	  \hline
	   Type                               & $\epsilon[\text{IoU}]$ (\%) &   $\epsilon[\text{PE}]$ (\%)\\
	  \hline
	  RaC \cite{general3d2020}            &                     76.29  &                        8.07  \\
	  NonCuboid \cite{NonCuboidRoom2022}  &                     81.40  &                        5.87  \\
	  HGC                                 &             \textbf{94.66} &                \textbf{2.57} \\
	  HGC \tnote{+}                       &                      50.76 &                        41.27 \\
	  \hline
	  \end{tabular}
	\begin{tablenotes}
     \item [+] Joint Wireframe Detection.
  \end{tablenotes}
 \end{threeparttable}
\end{table}

\subsection{Qualitative Comparison of Models}
\begin{figure*}[!t]
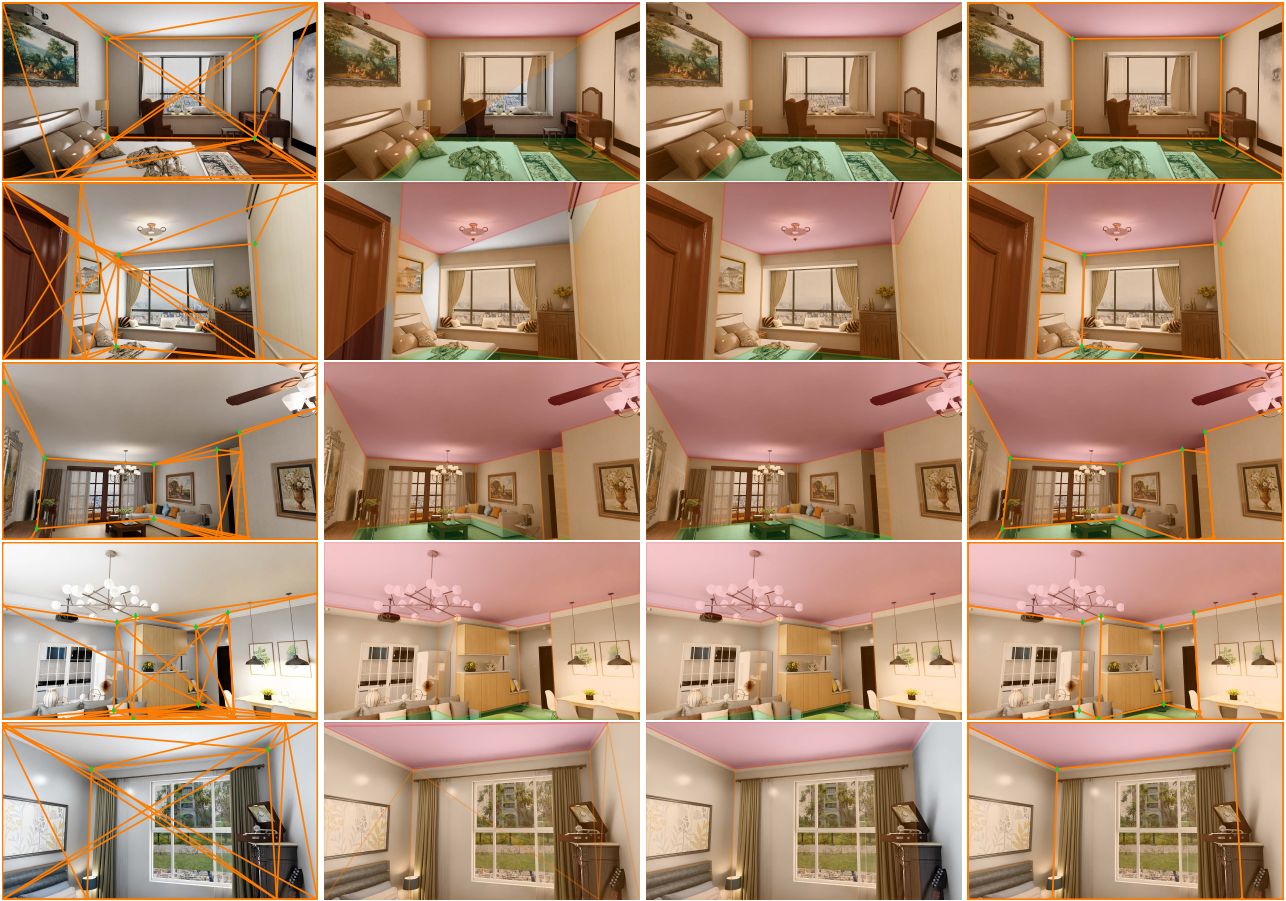

	\centering
	\exampleImageSynthetic{S03279R00525P0}\\
	\exampleImageSynthetic{S03279R00638P4}\\
	\exampleImageSynthetic{S03279R01340P3}\\
	\exampleImageSynthetic{S03337R02413P0}\\
	\exampleImageSynthetic{S03337R676922P2}\\
  \caption{Example output from the CSP and HGC model on simulated wireframe detections.
  From the left (i) Ground truth with sampled wireframe on.
  (ii) CSP estimate. (iii) HGC estimate. (iv) Ground truth polygons.}
  \label{fig:example_gt_wireframe}
\end{figure*}

Finally, we have a qualitative comparison between the models tested in this paper.
In Figure \ref{fig:example_gt_wireframe} we see semantic polygon detections from synthetic wireframes.
The first column shows the wireframe given as input to the model, the second column is the output of the Cycle Sampling Polygon-based classifier with NMS,
the third column is the Heterogeneous Graph-based Classifier with NMS and finally column four holds the ground truth annotation.

\begin{figure*}[!t]
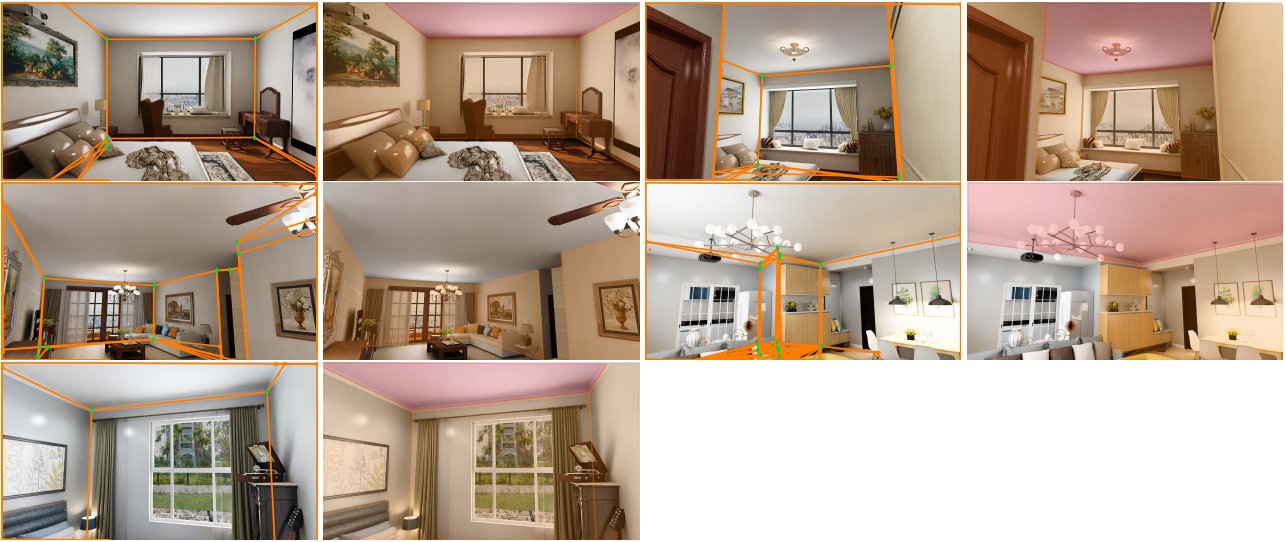

	\raggedright
	\exampleImageJoint{S03279R00525P0}%
	\exampleImageJoint{S03279R00638P4}\\%
	\exampleImageJoint{S03279R01340P3}%
	\exampleImageJoint{S03337R02413P0}\\%
	\exampleImageJoint{S03337R676922P2}\\%
  \caption{Example output from the HGC model when jointly detecting wireframes.
  Images are alternating (i) Ground truth with wireframe detections on, then (ii) HGC estimate.}
	\label{fig:example_det_wireframe}
\end{figure*}

In Figure \ref{fig:example_det_wireframe} we see similar images but for the HGC model with joint Wireframe detection and NMS.
Here images are alternating in pairs with the first being the detected wireframe and the second being the polygon output of the model.

\renewcommand{\figwidth}{0.24\linewidth}
\renewcommand{\hatchsmall}{_nohatch}
\renewcommand{\hatchlarge}{_NOHATCH}

\section{Conclusion}
In this paper we introduce a new neural network based system for semantic plane detection. The system takes a single image as input and outputs junctions, lines and planes corresponding to room layout estimation. The method uses an end-to-end learnt feature space, that is used both for detection and feature representations of junctions, lines and planes. These are then represented as a heterogeneous graph. This representation is then refined using two heterogeneous graph transformers followed by a classification network to provide the final output.
The methods are evaluated on the Structured3D dataset. The experiments show that given sufficiently good wireframe detections the model can outperform the state of the art Room Layout estimation models on 2D metrics and a promising direction for research.

\cleardoublepage

\section*{Acknowledgment}
This work was partially supported by the strategic research projects ELLIIT and eSSENCE, the Swedish Foundation for
Strategic Research project, Semantic Mapping and Visual Navigation for Smart Robots (grant no. RIT15-0038) and
Wallenberg Artiﬁcial Intelligence, Autonomous Systems and Software Program (WASP) funded by Knut and Alice Wallenberg Foundation.
The model training was enabled by the supercomputing resource Berzelius provided by National Supercomputer Centre at Linköping University
and the Knut and Alice Wallenberg foundation.

{\small
\bibliographystyle{ieee_fullname}
\bibliography{ref}
}
\cleardoublepage
\setcounter{section}{0}
\renewcommand*{\thesection}{A-\arabic{section}}
\section{Introduction}
This is supplementary material for the paper "Polygon Detection for Room Layout Estimation using Heterogenous Graphs and Wireframes".
It includes a more thorough explanation of the cycle generating algorithm, the maximum average weight cycle algorithm and statistics for inference time.

\section{Cycle Generation}
The CSP model described in section \ref{sec:CSP} generated polygons from the wireframe detections which are later classified.
This section describes how the polygons are generated.
Take the wireframe junctions and lines and form a homogeneous undirected graph $\GG$  where each vertex is a junction and each edge is a line.
In short we first find all cycles, then we check for uniqueness, and finally we make sure each cycle is a polygon without self-intersections.
From $\GG$ we find all connected subgraphs ${\Scal} = \{\GG_s | \GG_s \in \GG \}$, which are disjoint.
For each subgraph $\GG_s$ we find the cycle basis $\Cbf_s = \{C_1, C_2, ... C_Q\}$, consisting of cycles.
With the cycle basis it is possible to generate a cycle \cite{lee1982algorithmic} in $\GG_s$ by taking any connected subset $\Theta_k \subseteq \Cbf_s$
such that each $C_i \in \Theta_k$ is connected to at least one of the other cycles, i.e
\begin{equation}
	 C_i \cap \left ( \Theta_k \setminus C_i \right ) \neq \emptyset.
\end{equation}
The cycle is formed by taking the XOR (exclusive disjunction) product of all base cycles in $C_i \in \Theta_k$. Let
\begin{equation}
	 Z = X \oplus Y
\end{equation}
denote the XOR product between the two graphs $X$ and $Y$. Then the vertices $V_z$ of $Z$ will
be the cartesian product of the vertices $V_X, V_Y$ from $X$ and $Y$ respectively, i.e. $V_Z  = V_X \times V_Y$.
For all vertex pairs $(v_i, v_j) \in V_Z$ we form an edge $e = (v_i,v_j)$ in $Z$ if $e \in X$ or $e \in Y$ but not if $e \in X \cap Y$.
From the chosen basis subset $\Theta_k = \{ C_1, C_2, ..., C_R \}$ we form a new cycle
\begin{equation}
	 C_k = C_1 \oplus C_2 \oplus ... \oplus C_R
\end{equation}
and check if the generated cycle $C_k$ has at least 3 vertices and is a geometric valid polygon without intersections.
If so we say that $C_s$ is a valid polygon. To generate all polygons this is iterated for
all permutations of connected subsets in each connected subgraph $\Cbf_s$.

Calculating the cycle bases is cheap compared to generating all cycles, so this approach saves time during training since we
only generate a fixed amount of polygons during training. For inference however we must find all polygons, which does not scale well.

\section{Maximum Average Weight Cycle}
For the HGC model described in section \ref{sec:HGC} we need to find the best cycle from the neural networks edge scoring in the proposal step.
We want to find the cycle $C^*$ in a homogeneous undirected graph $\GG$ with the maximum average edge weight.
While there are many methods for finding the minimum average edge weight cycle in a directed graph, for example the algorithm by Karp \cite{karp},
we did not manage to restrict it to cycles with at least three edges.
Which is a problem since the optimal cycle will be traversing the highest scoring edge twice and be done.
Therefore we use a greedy method based on the shortest path algorithm of Djikstra \cite{Dijkstra}.

For each plane anchor $A_k$ we form a graph $\GG_k$ were each edge $e_j \in \GG_k$ correspond to detected line $L_j$ and each
vertex $v_i \in \GG_k$ correspond to an junction $J_i$.
For an edge $e_j$ we calculate a score
\begin{equation}
	s_j = \sigma \left ( \gbfh_j^T W \ddh_k \right ),
\end{equation}
as explained in section \ref{sec:proposal_network}.
From the score each edge is given a weight $w_i=1-s_i$ to formulate the problem as finding the minimum average weight cycle.

The algorithm is outlined in Algorithm \ref{alg:min_cycle} and will iteratively try a different edge $e_s \in \GG_k$ as starting point for
minimum weight cycles. The cycle is found by taking the vertices of the edge as start respectively target of the shortest path problem on the graph $\GG_k \setminus e_s$.
By finding the shortest path and adding $e_s$ we have a minimum weight cycle containing the starting edge.
Edges are tried as starting edge going from lowest to highest weight for a fixed amount of iterations $T$.

\begin{algorithm}
	\SetKwFunction{shortestPath}{shortestPath}
	\SetKwFunction{sort}{sort}
	\SetKwFunction{takeMinWeightEdge}{takeMinWeightEdge}
	\SetKwFunction{isPolygon}{isValidPolygon}
	\SetKwFunction{numberOfEdges}{numberOfEdges}
	\SetKw{and}{and}
	\SetAlgoLined
	\KwData{
		Graph : $\GG_k$, \\
		Edges and weights: $(e_j, w_j) \in \GG_k$, \\
		Iterations: $T$ \\
	}
	\KwResult{Polygon $C^*$ with approximate lowest average weight $w^*$}
	$E := \{e_j \; | \; (e_j, w_j) \in \GG_k \}$, candidate edges \\
	$w^* := \infty$ \\
	$C^* := \emptyset$ \\
	\For{$t = 1,...,T$}{
		$e_t :=$ \takeMinWeightEdge{$E$} \\
		$E := E \setminus e_s$ \\
		$\GG_t$ := $\GG_k \setminus e_t$ \\
		$C_t :=$ \shortestPath{$\GG_t, e_t[1], e_t[2]$} \\
		$w_t := \text{mean}(\{w_j \; | \; (e_j, w_j) \in C_t \} )$ \\
		\If{$w_t < w^*$ \and \numberOfEdges{$C_t$} $\ge 3$ \and \isPolygon{$C_t$}}{
			$w^* := w_t$ \\
			$C^* := C_t$ \\
		}
	}
	\caption{An overview of how the algorithm finds an approximate minimum average weight polygon.}
	\label{alg:min_cycle}
\end{algorithm}

\section{Inference Time}
We measure inference time by evaluating the model over 1000 images in the validation set on an NVidia Titan V GPU.
Measuring from having the image in RAM to getting the result back on the GPU.
In table \ref{tab:timings} we see that the Graph-based HGC method is much faster on average while also
having much smaller variations in inference time. See boxplots in Figure \ref{fig:timings} for a visualization.

\begin{table}
	\begin{threeparttable}[b]
	  \caption{Median and mean inference times in seconds with standard deviation for the CSP and HGC model.}
	  \label{tab:timings}
	  \centering
	  \begin{tabular}{|l|r|r|r|r|r|}
	  \hline
	   Model           & Median  & Mean & Std \\
	  \hline
	  CSP              &  8.88   & 0.72 & 11.98 \\
	  HGC              &  0.10   & 0.11 & 0.06 \\
		\hline
	  HGC \tnote{+}    &  0.12   & 0.12 & 0.06 \\
	  \hline
	  \end{tabular}
		\begin{tablenotes}
       \item [+] Joint Wireframe Detection.
    \end{tablenotes}
	 \end{threeparttable}
\end{table}

\renewcommand{\figwidth}{0.93\linewidth}
\begin{figure}[!t]
	\centering
	\begin{subfigure}[b]{\figwidth}
         \centering
         \includegraphics[width=\figwidth, trim = 0.5cm 1cm 1cm 1.3cm, clip]{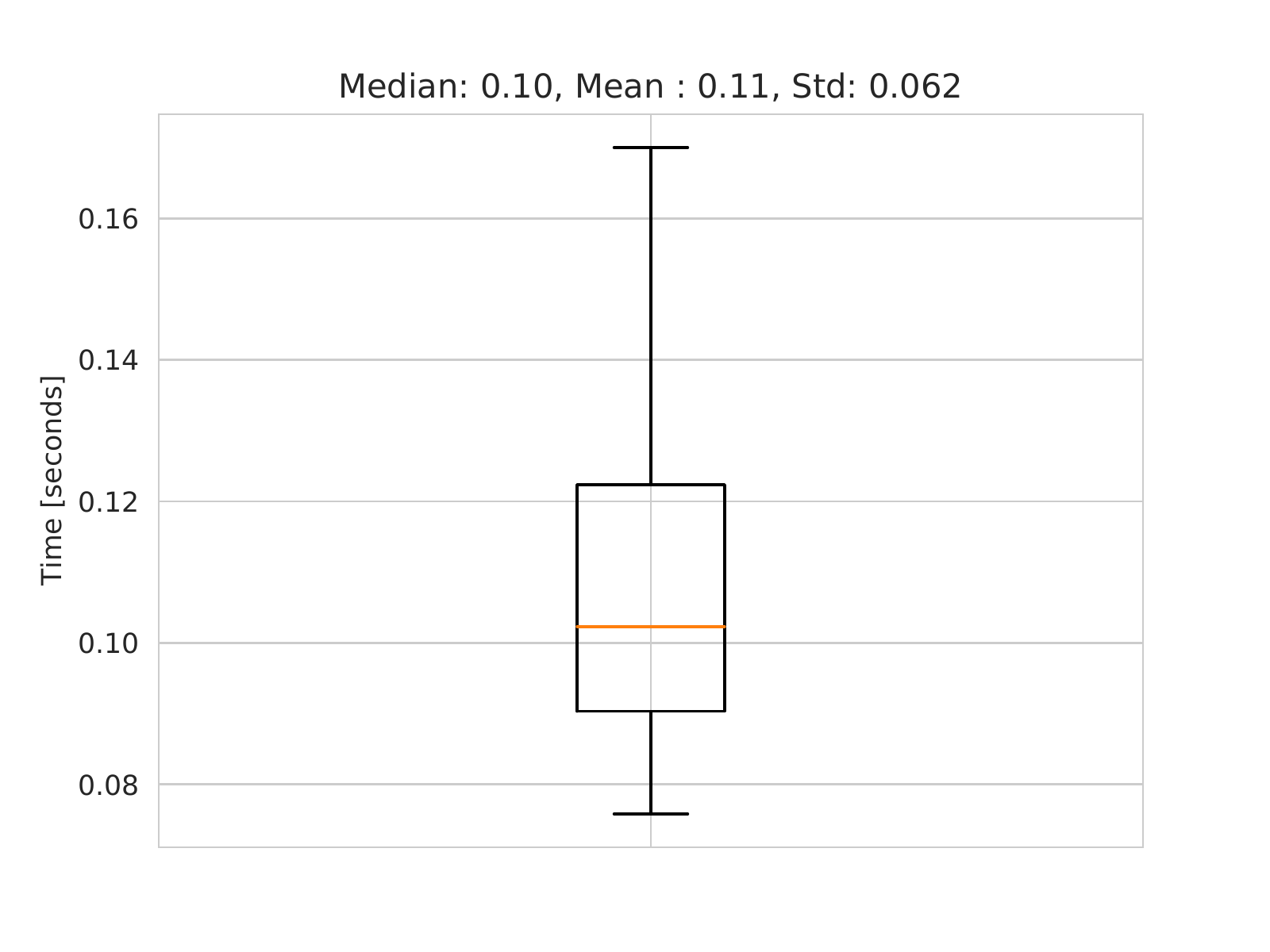}\\%
         \caption{Heterogeneous Graph model with synthetic wireframe.}
  \end{subfigure}\\%
	\begin{subfigure}[b]{\figwidth}
         \centering
         \includegraphics[width=\figwidth, trim = 0.5cm 1cm 1cm 1.3cm, clip]{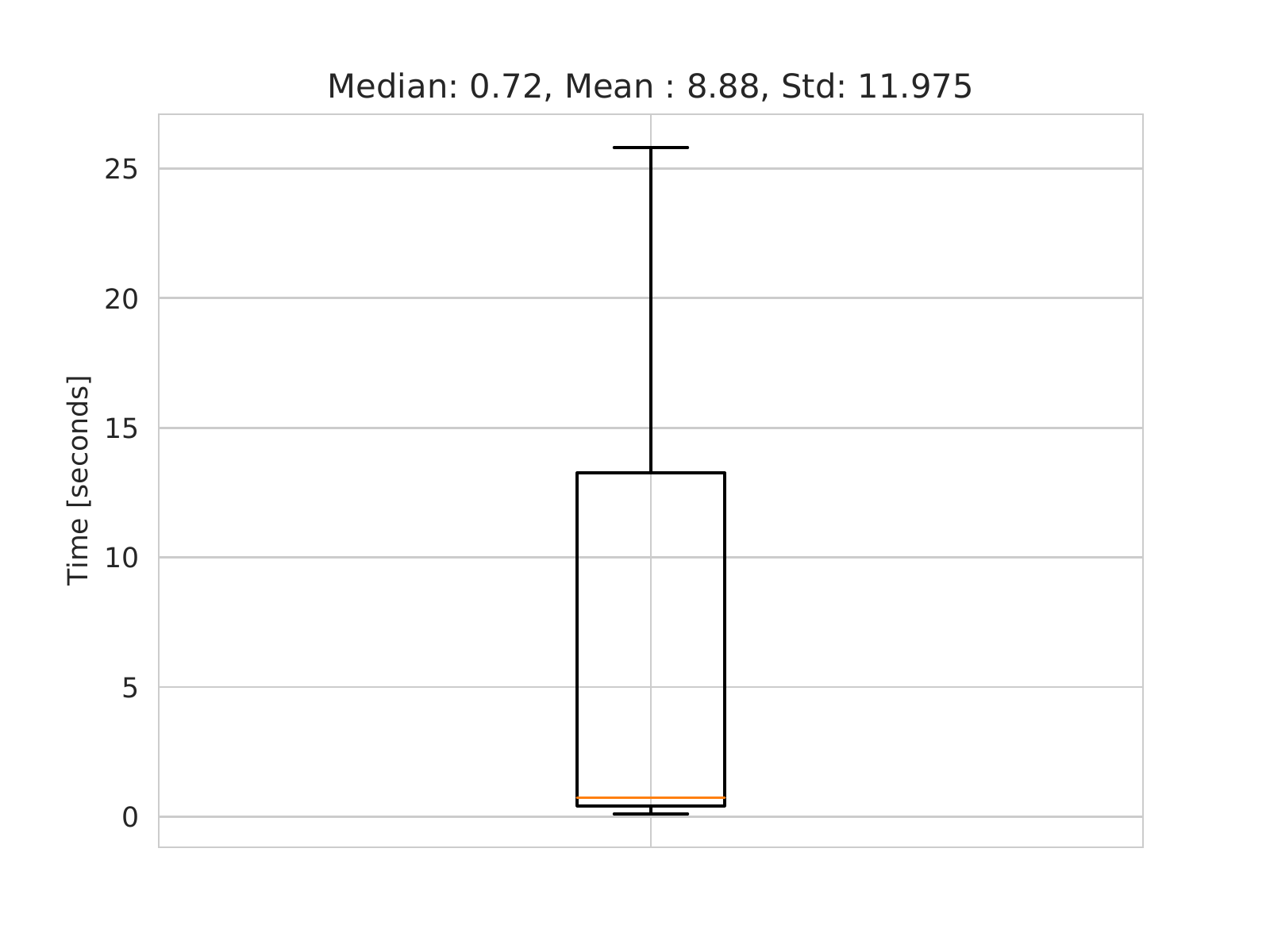}\\%
         \caption{Cycle Sampling based model with synthetic wireframe.}
  \end{subfigure}\\%
	\begin{subfigure}[b]{\figwidth}
         \centering
         \includegraphics[width=\figwidth, trim = 0.5cm 1cm 1cm 1.3cm, clip]{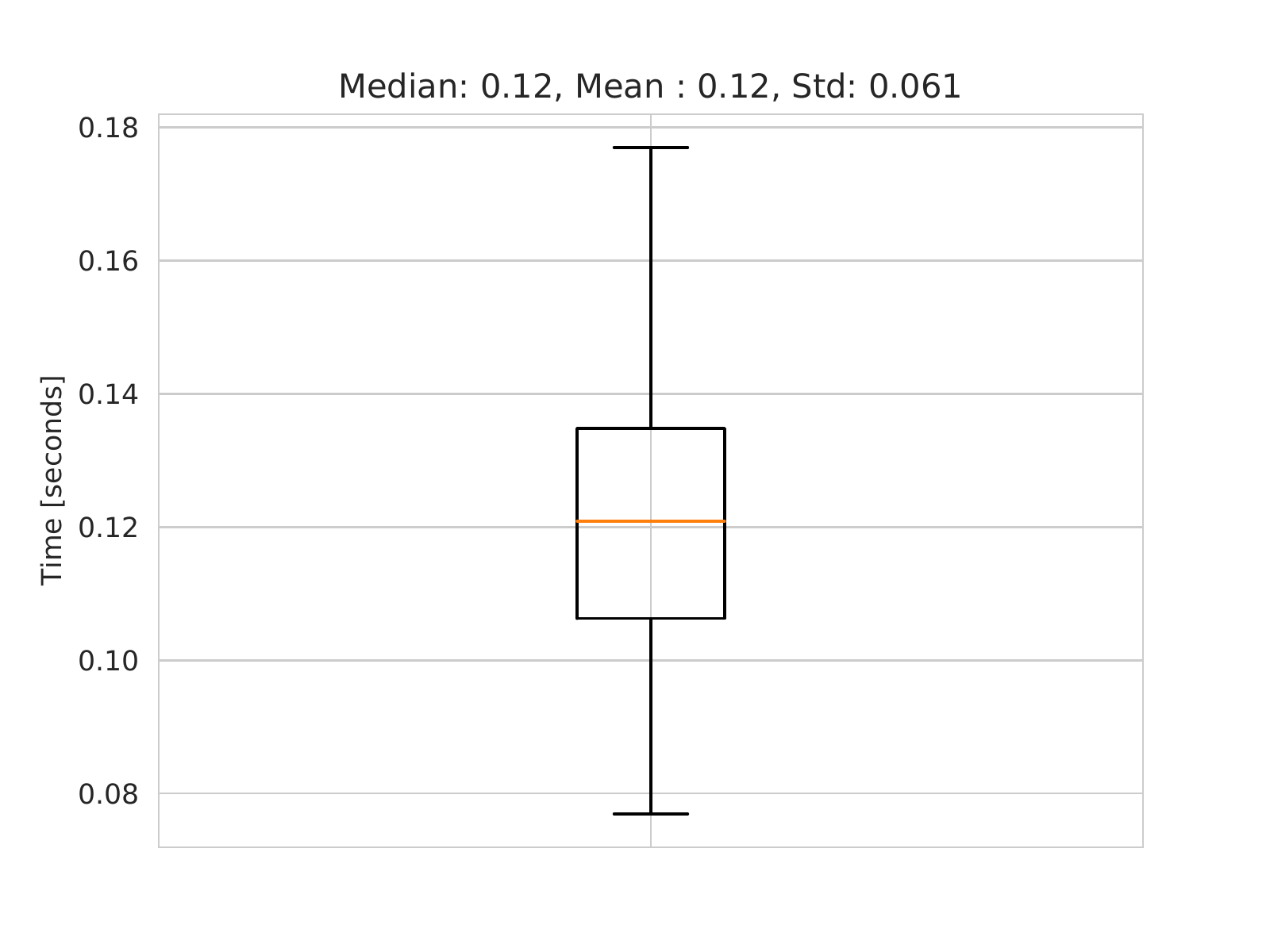}\\%
         \caption{Heterogeneous Graph model with predicted wireframe.}
  \end{subfigure}\\%
  \caption{Inference times for the two different models. HCG is tested with synthetic and joint wireframe prediction.}
  \label{fig:timings}
\end{figure}

\end{document}